\documentclass[10pt,twocolumn,letterpaper]{article}

\usepackage[pagenumbers]{cvpr} 
\usepackage[numbers,sort&compress]{natbib}

\usepackage{graphicx}
\graphicspath{{assets/}} 
\usepackage{amsmath}
\usepackage{amssymb}
\usepackage{booktabs}
\usepackage{enumitem}
\usepackage{algorithm}
\usepackage{algpseudocode}
\usepackage{bm}
\usepackage{multirow}
\usepackage{caption}
\usepackage{subcaption}
\usepackage{tikz}

\usepackage{pifont}
\newcommand{\cmark}{\ding{51}}%
\newcommand{\xmark}{\ding{55}}%

\usepackage{lipsum}
\usepackage{bbm}
\usepackage{dsfont}
\usepackage{listings}


\definecolor{cvprblue}{rgb}{0.21,0.49,0.74}
\usepackage[pagebackref,breaklinks,colorlinks,citecolor=cvprblue]{hyperref}

\usepackage[capitalize]{cleveref}
\crefname{section}{Sec.}{Secs.}
\Crefname{section}{Section}{Sections}
\Crefname{table}{Table}{Tables}
\crefname{table}{Tab.}{Tabs.}
\newcommand{\negtome}{NegToMe\xspace}
\def\paperID{****} 
\def\confName{CVPR}
\def\confYear{2025}

\begin{document}

\title{Negative Token Merging: Image-based Adversarial Feature Guidance}

\definecolor{alphaColor}{HTML}{FF9100} 
\definecolor{betaColor}{HTML}{00D5FF} 
\definecolor{gammaColor}{HTML}{F0539B} 

\newcommand{\instalpha}{{\color{alphaColor}{\alpha}}}
\newcommand{\instbeta}{{\color{betaColor}{\beta}}}
\newcommand{\instgamma}{{\color{gammaColor}{\chi}}}
\newcommand\swj[1]{\draftcomment{\textcolor{orange}{[\textit{#1}]$_{-\text{WS}}$}}}
\newcommand\lj[1]{\draftcomment{\textcolor{blue}{[\textit{#1}]$_{-\text{Lindsey}}$}}}



\author{
Jaskirat Singh$^{\instalpha*}$ \quad Lindsey Li$^{\instbeta*}$ \quad Weijia Shi$^{\instbeta*}$ \quad Ranjay Krishna$^{\instbeta \instgamma}$ \quad Yejin Choi$^{\instbeta}$ \\
Pang Wei Koh$^{\instbeta \instgamma}$ \quad Michael F. Cohen$^{\instbeta}$ \quad Stephen Gould$^{\instalpha}$ \quad Liang Zheng$^{\instalpha}$ \quad Luke Zettlemoyer$^{\instbeta}$
\\
 $^{\instbeta}$University of Washington \quad  $^{\instalpha}$Australian National University \quad $^{\instgamma}$Allen Institute for AI\\
}



\twocolumn[{
\maketitle
\begin{center}
    \vskip -0.25in
    \captionsetup{type=figure}
    \begin{subfigure}[b]{1.0\textwidth}
         \centering
         \includegraphics[width=\textwidth]{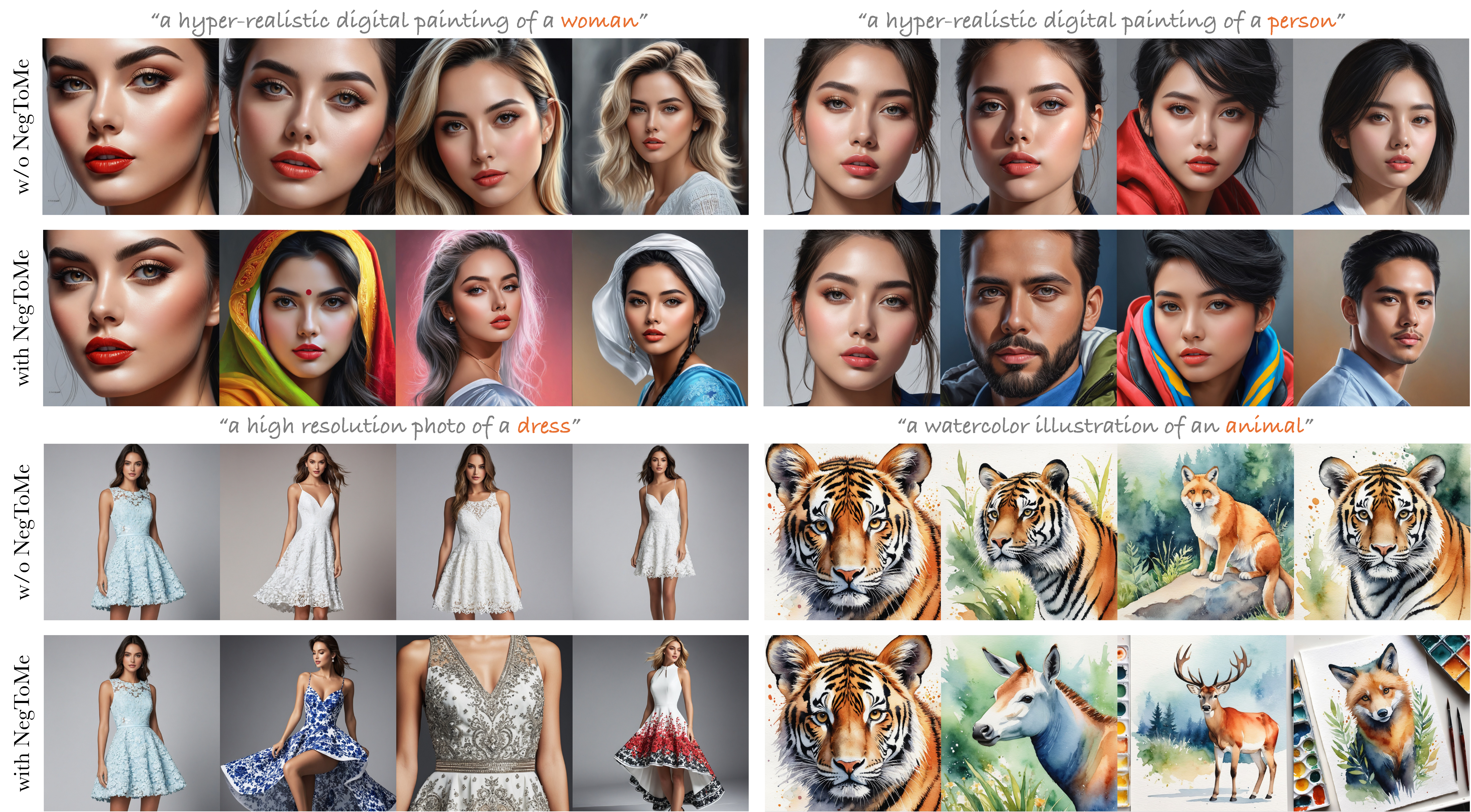}
         \caption{\textbf{Adversarial Guidance across Different Outputs:} State-of-the-art diffusion models are observed to suffer from limited diversity (\eg, ethnic, racial, gender \etc). NegToMe can be used to improve output diversity by simply guiding the features of each image away from each other during reverse diffusion.
         }
     \end{subfigure} 
     \begin{subfigure}[b]{1.0\textwidth}
         \centering
         \includegraphics[width=\textwidth]{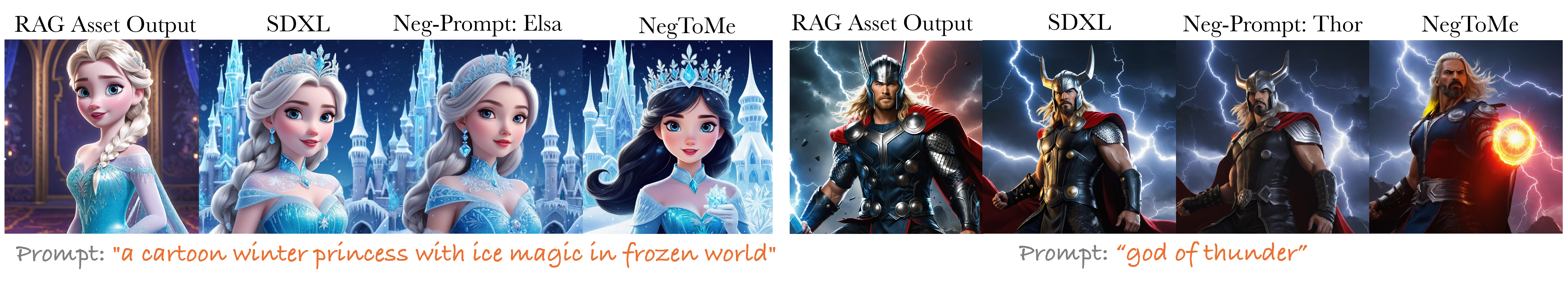}
         \vskip -0.1in
         \caption{{\textbf{Adversarial Guidance with Copyrighted Content:} Diffusion models can generate copyrighted content. Moreover, using negative prompt for avoiding this is often insufficient. NegToMe helps better reduce similarity to copyrighted characters, by guiding diffusion features away from copyrighted images}.}
     \end{subfigure}
    \vskip -0.05in
    \caption{
    We introduce \negtome, a 
    training-free approach for adversarial guidance directly using images instead of a negative prompt.
    Above we show its applications for a) improving output diversity (visual, gender, racial) by guiding each image away from others, b) reducing visual similarity to copyrighted characters, by guiding outputs away from copyrighted images. (refer Sec.~\ref{sec:experiments} for further applications). 
   }
     \label{fig:overview}
\end{center}
}
]

\begin{abstract}
    \vskip -0.15in

Text-based adversarial guidance using a negative prompt has emerged as a widely adopted approach to steer diffusion models away from producing undesired concepts.
While useful, performing adversarial guidance using text alone can be insufficient to capture complex visual concepts or avoid specific visual elements like copyrighted characters. 
In this paper, for the first time we explore an alternate modality in this direction by performing adversarial guidance directly using visual features from a reference image or other images in a batch. We introduce \textbf{neg}ative \textbf{to}ken \textbf{me}rging (\negtome), a simple but effective training-free approach which performs adversarial guidance through images by  selectively pushing apart matching visual features between reference and generated images during the reverse diffusion process. 
By simply adjusting the used reference, NegToMe enables a diverse range of applications. Notably, when using other images in same batch as reference, we find that \negtome significantly enhances output diversity (e.g., racial, gender, visual) by guiding features of each image away from others.
Similarly, when used w.r.t.~copyrighted reference images, \negtome reduces visual similarity to copyrighted content by 34.57\%. \negtome is simple to implement using just few-lines of code, uses only marginally higher ($<4\%$) inference time and is compatible with different 
    diffusion architectures, including those like Flux, which don't natively support the use of a negative prompt. Code is available at \url{https://negtome.github.io}.
    \vskip -0.25in
\end{abstract}

\section{Introduction}
\label{sec:intro}

\begin{figure}[t]
\vskip -0.25in
\begin{center}
\centerline{\includegraphics[width=1.\linewidth]{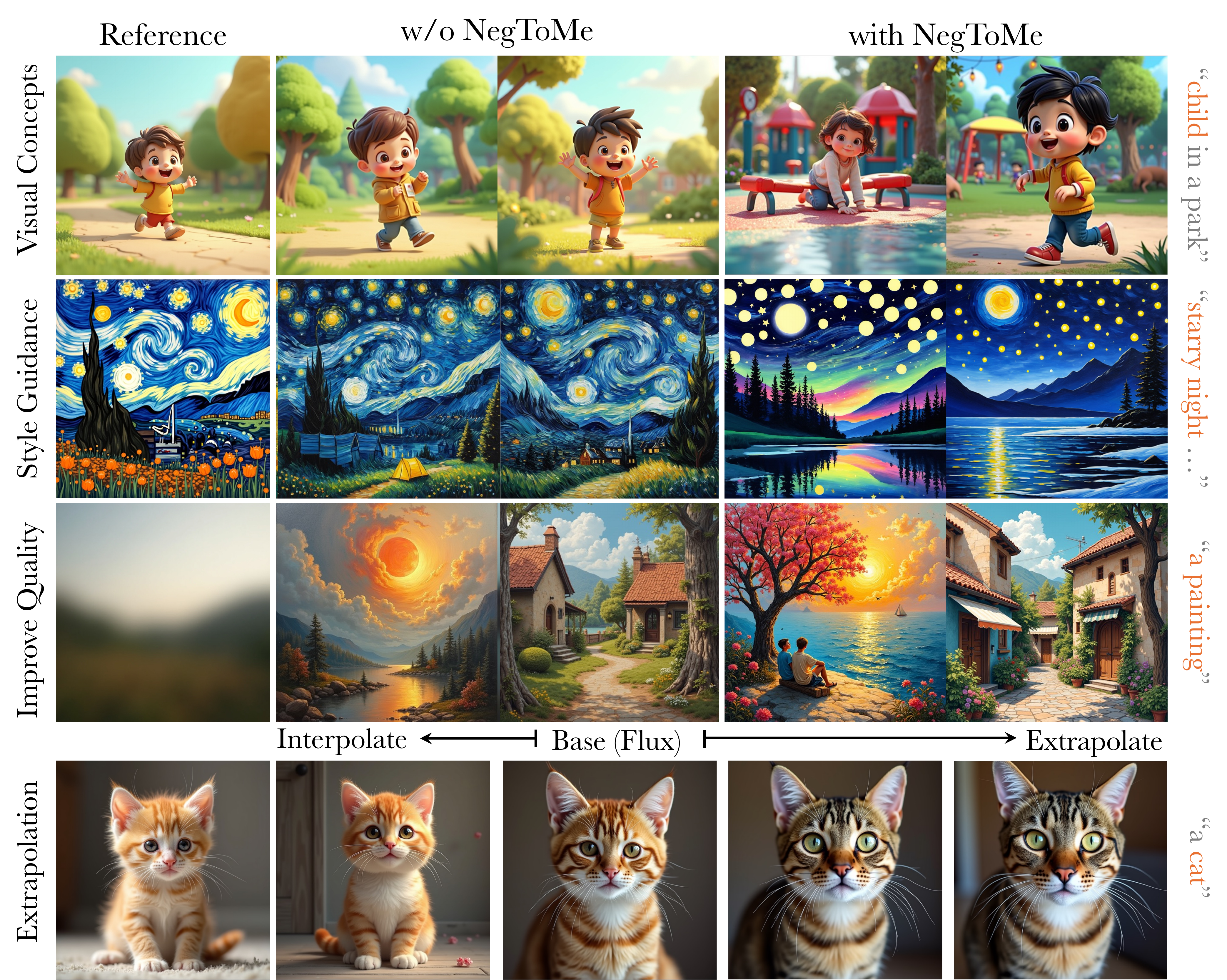}}
\vskip -0.1in
\caption{
\emph{\textbf{Image-based adversarial guidance.}} \negtome enables directly using images (instead of negative prompt alone) for adversarial guidance. By simply adjusting the used reference, NegToMe allows for a range of custom applications, 1) adversarial guidance for visually complex concepts to improve diversity, 2)  Style control for excluding specific artistic styles (\eg, Van Gogh) from generated images, 3) improving output quality by guiding away from a blurry reference, 4) Object feature interpolation or extrapolation by guiding the outputs towards or away from the reference \etc.
}
\label{fig:general-use}
\end{center}
\vskip -0.45in
\end{figure}

Large-scale text-to-image (T2I) diffusion models \cite{rombach2021highresolution,nichol2021glide,saharia2022photorealistic,ramesh2022hierarchical,yu2022scaling, dai2023emu} have made unparalleled progress and allow for generation of powerful imagery. 
Despite these advances, guiding the generation process adversarially to avoid generation of undesired concepts \cite{ban2024understanding} remains a challenging problem. 
Such guidance is advantageous for several applications such as improving image quality (by guiding away from low-quality features), improving output diversity (by guiding each image away from each other), avoiding undesired concepts such as copyrighted characters (Fig.~\ref{fig:overview}, \ref{fig:general-use}) \cite{he2024fantastic} \etc.

Existing methods in this direction predominantly rely on the use of negative prompt \cite{ho2022classifier, armandpour2023re} for adversarial guidance.
However, use of negative-prompt \emph{alone} for adversarial guidance suffers from some limitations; capturing complex visual concepts using text-alone can be hard
(trying to capture every detail: pose, action, background \etc for \emph{child in a park} in Fig.~\ref{fig:general-use}). The use of negative-prompt \emph{alone} might be insufficient to remove undesirable visual features (\eg, copyrighted characters in Fig.~\ref{fig:overview}). Furthermore, using a separate negative prompt itself may not be feasible when using \emph{state-of-the-art} guidance distilled models like Flux \cite{flux2024}.


To address this challenge, we propose \emph{\textbf{neg}ative \textbf{to}ken \textbf{me}rging} (\negtome), a simple training-free approach that performs adversarial guidance directly using \textit{images} instead of text.  
Our key insight is that even when text is not sufficient to capture visually complex concepts (\emph{child in park} for Fig.~\ref{fig:general-use}), we can directly use the visual features from a reference image to adversarially guide the generation process.  For instance in Fig.~\ref{fig:general-use}, instead of trying to exhaustively describe the \emph{child's} attire, placement, pose, background \etc,  we can directly use the reference image for better guidance. Similarly, in cases where a negative prompt alone is not sufficient (\eg, copyrighted characters in Fig.~\ref{fig:overview}), we can better guide the generation away from undesired concepts by directly using the character images for adversarial guidance.

\negtome is easy to implement and can be incorporated into various diffusion architectures using only a few lines of code (Alg.~\ref{alg:negtome}). By simply adjusting the used reference, NegToMe enables a range of custom applications
(see Fig.~\ref{fig:general-use}); 1) adversarial guidance for visually complex concepts (\eg, \emph{`child in a park'}), 2) style-guidance for excluding specific artistic elements, 3) enhancing output aesthetics by guiding away from a blurry reference, and, 4) object feature interpolation or extrapolation by guiding the outputs towards or away from the reference \eg, between a \emph{kitten} and \emph{young cat} (Fig.~\ref{fig:general-use})
to inter-extrapolate features for cat age, size \etc.

In particular, to demonstrate the practical usefulness of the proposed approach, we identify two prominent use-cases of \negtome; 1) improving output diversity 
(when using other images in same batch as reference), 
and, 2)  reducing visual similarity to copyrighted characters (when performed \emph{w.r.t} a copyrighted RAG database).
For instance, it has been empirically shown \cite{miao2024training} that state-of-the-art diffusion models often struggle from the problem of limited output diversity (\eg, limited racial, gender diversity for a prompt for a \emph{person} in Fig.~\ref{fig:overview}).  
The use of \negtome across a batch, inherently helps addresses this problem by pushing visual features of each image away from the each other during reverse diffusion process.
Through both qualitative and quantitative results (refer Sec.~\ref{sec:output-diversity}) we show that \negtome helps significantly improve output diversity (racial, gender, ethnic, visual) without requiring any training or finetuning.

\begin{figure*}[t]
\vskip -0.35in
\begin{center}
\centerline{\includegraphics[width=1.0\linewidth]{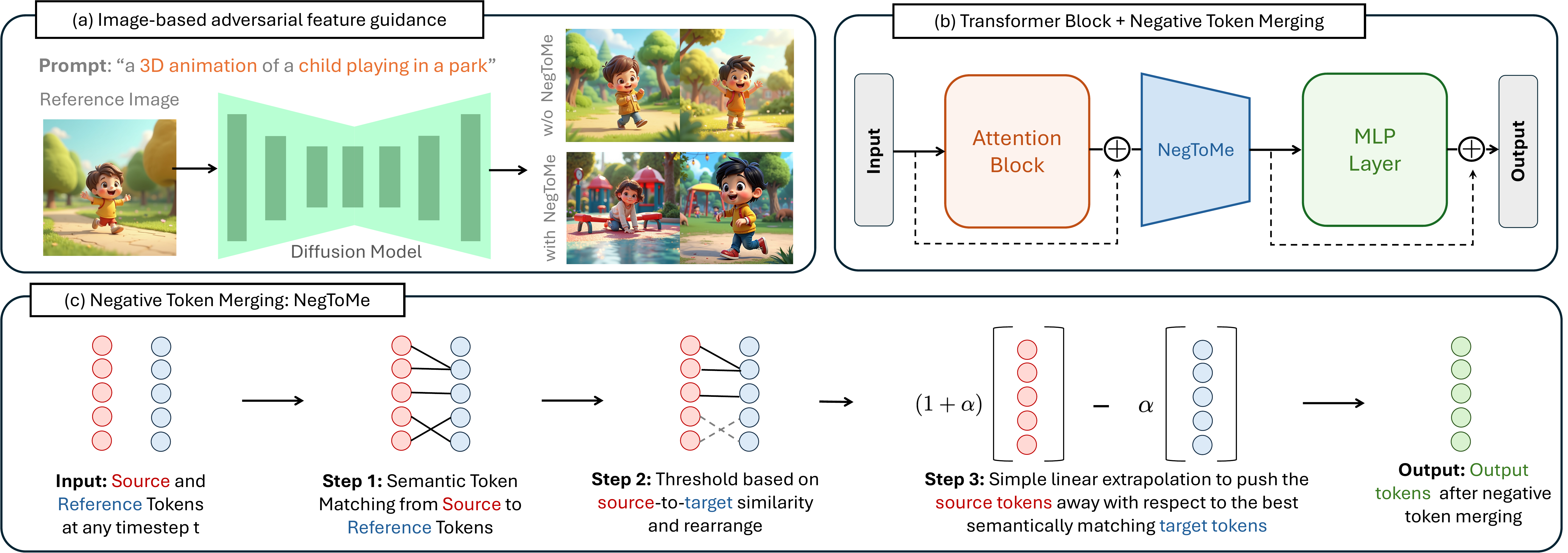}}
\vskip -0.1in
\caption{\emph{\textbf{Method Overview.}} (a) The core idea of NegToMe is to perform adversarial guidance directly using visual features from a reference image (or other images in the same batch). (b) NegToMe is simple and can be applied in any transformer block. (c) A simple three step process for performing adversarial guidance using NegToMe (refer Sec.~\ref{sec:method} and Alg.~\ref{alg:negtome} for the detailed implementation).}
\label{fig:method-overview}
\end{center}
\vskip -0.3in
\end{figure*}

Similarly, when performed \emph{w.r.t} to a copyrighted RAG database (Sec.~\ref{sec:copyright}), we observe that \negtome complements and improves over traditional negative-prompt based adversarial guidance \cite{he2024fantastic} for reducing visual similarity with copyrighted characters. Experiments reveals that despite its simplicity, the proposed approach helps reduce visual similarity to copyrighted content by 34.57\% while using only marginally higher ($<4\%$) inference times (Sec.~\ref{sec:experiments}).

\section{Related work}

\textbf{Adversarial feature guidance} 
using a negative prompt has been widely explored for a range of applications \cite{ho2022classifier, ban2024understanding, he2024fantastic, andrew2023negativeprompts, woolf2023negativeprompts}.  While remarkable, as discussed in Sec.~\ref{sec:intro} we find that the use of a negative-prompt alone might not always be sufficient (Fig.~\ref{fig:overview}, \ref{fig:general-use}) or even sometimes feasible \cite{flux2024}. Our work thus explores a complementary modality by performing adversarial guidance directly using 
a reference image.

\noindent\textbf{Token merging} \citep{bolya2022token, bolya2023tomesd} proposes to increase the throughput of existing ViT \cite{dosovitskiy2020image} models by gradually merging redundant tokens in the transformer blocks. 
Recent works \citep{li2024vidtome, wu2024fairy} apply the idea of token merging for video editing in order to better maintain temporal coherence of the edited video. In contrast, we explore cross-frame negative token merging as a mechanism for providing adversarial feature guidance. \

\noindent\textbf{Increasing output diversity} has been explored to address mode-collapse with 
diffusion models \cite{qin2023class, zhang2023iti, bansal2022well, miao2024training}. The reduced diversity occurs due a several factors including training-data imbalance, classifier-free guidance \cite{ho2022classifier}, preference-optimization finetuning \cite{rafailov2024direct, lee2023aligning} \etc, and is hard to eliminate at pretraining stage. Prior works on addressing this often require costly retraining /  finetuning \cite{miao2024training}. In contrast, we propose a simple training-free approach which inherently improves semantic diversity of the output features.

\noindent\textbf{Copyright mitigation}. 
The growing concern over copyright infringement by generative models, 
has attracted significant attention in recent literature \cite{sag2018new, Henderson2023FoundationMA, lee2024talkin, sag2023copyright, min2023silo, shi2024detecting}.
A particularly pressing issue is generation of copyrighted characters by diffusion models \cite{he2024fantastic, Henderson2023FoundationMA, lee2023talkin, golatkar2024cpr, wei2024evaluatingcopyrighttakedownmethods}.
Prior works for addressing these risks typically require expensive finetuning and unlearning \cite{gong2024reliableefficientconcepterasure, chefer2023attend, zhang2024forget} to remove copyrighted information from model weights. Our work (\negtome) thus provides a simple approach for reducing visual similarity to copyrighted content in a training-free manner.

\section{Negative Token Merging}
\label{sec:method}

Our goal is to insert a negative token merging module into an existing diffusion model \cite{podell2023sdxl, flux2024}, in order to perform adversarial feature guidance \emph{w.r.t} other images in a batch (Sec.~\ref{sec:output-diversity}) or a real-image input (Sec.~\ref{sec:copyright}). The core idea of our approach is to perform adversarial guidance by pushing each output token (\emph{source}) away from its best matching token (\emph{target}) in the reference image. The negative token merging module is applied between the attention and MLP branches of each transformer block (Fig.~\ref{fig:method-overview}). In particular, given the output of the attention block, we first perform cross-image token matching to find the best matching \emph{target} token for each output \emph{source} token. We then apply simple linear extrapolation pushing each \emph{source} token away from its best matching \emph{target} token. Fig.~\ref{fig:method-overview} provides an overview.

\textbf{Semantic Token Matching.}
A key idea behind \negtome is to push each output token (\emph{source}) away from its semantically closest token
(\emph{target}) in the reference image during the reverse diffusion process.
This requires computation of semantic token-to-token correspondences 
between the generated tokens and the tokens in the reference image. Luckily, we can leverage the rich semantic structure of intermediate diffusion features \cite{tang2023emergent, zhang2024tale} to compute cross-image token-token similarities using noisy features itself.

In particular, given the output $\mathrm{O}_{src} \in \mathbb{R}^{B \times N \times D}$ of an attention block ($B$ is the batch size and $N$ is the number of image tokens), we first compute similarity \emph{w.r.t} the reference image tokens $\mathrm{O}_{ref} \in \mathbb{R}^{1 \times N \times D}$  as follows,
\begin{align}
    \mathcal{S}(\mathrm{O}_{src}, \mathrm{O}_{ref}) = \Tilde{\mathrm{O}}_{src} \cdot \Tilde{\mathrm{O}}_{ref}^T; \quad   \mathcal{S} \in \mathbb{R}^{B.N \times N},
\end{align}
where $\Tilde{\mathrm{O}}_{src}, \ \Tilde{\mathrm{O}}_{ref}$ refer to the frame-level normalized source and reference image tokens, respectively. We next use the similarity matrix $\mathcal{S} \in \mathbb{R}^{B.N \times N}$ in order to compute the best matching target token for each source token as,
\begin{align}
    \mathrm{O}_{target} = \mathrm{O}_{ref} \ [\mathrm{argmax} \left
    \{\mathcal{S}(\mathrm{O}_{src}, \mathrm{O}_{ref}) \right\}]\\
    \mathrm{O}_{{target}} = \mathrm{H} \odot \mathrm{O}_{target} + (1 - \mathrm{H}) \odot \mathrm{O}_{src},
\end{align}
where $\mathrm{H} = \mathds{1}\left[  \mathrm{max} \{\{\mathcal{S}(\mathrm{O}_{src}, \mathrm{O}_{ref})\} > \tau  \right]$ helps ensure that \emph{source-tokens} with \emph{source-to-target} token similarity below a threshold $\tau$ (\ie, no good semantic match is available) are not modified during negative token merging.

\algrenewcommand\algorithmicindent{0.5em}%
\begin{figure}[t]
\vspace{-15pt}
\begin{algorithm}[H]
\caption{NegToMe: Negative Token Merging}
\label{alg:negtome}
\definecolor{codeblue}{rgb}{0.1,0.6,0.1}
\definecolor{codekw}{rgb}{0.85, 0.18, 0.50}
\lstset{
  backgroundcolor=\color{white},
  basicstyle=\fontsize{7.5pt}{7.5pt}\ttfamily\selectfont,
  columns=fullflexible,
  breaklines=true,
  captionpos=b,
  commentstyle=\fontsize{7.5pt}{7.5pt}\color{codeblue},
  keywordstyle=\fontsize{7.5pt}{7.5pt}\color{codekw},
  escapechar={|}, 
}
\vspace{-5pt}
\begin{lstlisting}[language=python]
def NegToMe(x_src, x_ref, alpha, threshold):
    """
    x_src: [B, N, D]
    x_ref: [N, D]
    alpha: float
    threshold: float
    """
    # 1) Normalization
    x_src_norm = F.normalize(x_src, dim=-1)
    x_ref_norm = F.normalize(x_ref, dim=-1)

    # 2) Cosine similarity
    cosine_similarity = x_src_norm @ x_ref_norm.T

    # 3) Find source-to-target match and rearrange
    max_similarity, argmax_indices = cosine_similarity.max(dim=-1)
    x_target = x_ref[argmax_indices]

    # 4) Threshold and merge
    threshold_mask = max_similarity > threshold
    x_merge = torch.where(
        threshold_mask.unsqueeze(-1),
        (1 + alpha) * x_src - alpha * x_target,
        x_src)
    return x_merge
\end{lstlisting}
\vspace{-5pt}
\label{alg1}
\end{algorithm}
\vspace{-20pt}
\end{figure}


\textbf{Source-to-Target Token Extrapolation.}
Given the semantically-matched \emph{target} token matrix for the \emph{source} tokens $\mathrm{O}_{src}$, we next perform a simple linear extrapolation between the \emph{source} and \emph{target} tokens as,
\begin{align}
    \mathrm{O}_{merge} =  (1 + \alpha_t)\  \mathrm{O}_{src} - \alpha_t \  \mathrm{O}_{target},
\label{eq:negative-token-mixing}
\end{align}
where $\alpha_t$ is a time-dependent affine-coefficient, which helps control the degree to which the \emph{source} and \emph{target} tokens are pushed apart during the reverse diffusion process.

\textbf{Masked Adversarial Guidance}. 
While performing adversarial guidance \emph{w.r.t} the entire reference image is useful (\eg, increasing diversity), we may also wish to perform adversarial guidance only \emph{w.r.t} to certain parts of the provided reference. For instance, when performing copyright mitigation (Sec.~\ref{sec:copyright}), we may wish to reduce visual similarity only with respect to the copyrighted character without being affected by background noise or features. Similarly, masked guidance may also be useful in order to perform adversarial guidance with specific subparts (\eg, red hat, mustache, pony tail) of the provided reference image.

In particular, given an additional mask binary input $\mathrm{M}_{ref}$ for the reference image, we can perform masked adversarial guidance by simply introducing a bias-term in the source-to-target similarity $\mathcal{S} \in \mathbb{R}^{B.N \times N}$ computation as,
\begin{align}
    \mathcal{S}(\mathrm{O}_{src}, \mathrm{O}_{ref}) = \Tilde{\mathrm{O}}_{src} \cdot \Tilde{\mathrm{O}}^T_{ref} + \log(\Tilde{M}_{ref} + \epsilon),
\end{align}
where $\epsilon=10^{-6}$ and $\Tilde{M}_{ref}\in\mathbb{R}^{1 \times N}$ is  original mask $\mathrm{M}_{ref}$ resized and flattened to match the corresponding sequence length $N$ for attention block output $\mathrm{O}_{src} \in \mathbb{R}^{B \times N \times D}$.

\textbf{Application to MM-DiT Models.}
A key advantage of the proposed approach is that it is also easily extendable to guidance-distilled MM-DiT architectures such as Flux \cite{flux2024}, which do not natively support the use of a separate negative prompt. In particular,  given an output $\mathrm{O}_{joint} \leftarrow (\mathrm{O}_{text}, \mathrm{O}_{img})$ of the joint-attention block \cite{flux2024,esser2024scaling}, NegToMe can be easily applied as follows,
\begin{align}
    \mathrm{O}_{joint} \leftarrow \{ \mathrm{O}_{text} \ \oplus \ f_{neg}(\mathrm{O}_{img},  \ \mathrm{O}_{ref}, \alpha_t, \tau) \},
\end{align}
where $f_{neg} (.)$ is NegToMe function from Alg.~\ref{alg:negtome}, and $\oplus$ is the matrix concatenation operation along sequence length.
\section{Experiments}
\label{sec:experiments}

In this section, we demonstrate the practical usefulness of our approach for two prominent applications of \negtome; increasing output diversity (Sec.~\ref{sec:output-diversity}) and reducing visual similarities to copyrighted characters  (Sec.~\ref{sec:copyright}). We further showcase more general applications of \negtome in Sec.~\ref{sec:analysis}.

\subsection{Increasing Output Diversity}
\label{sec:output-diversity}

\begin{figure*}[t]
\vskip -0.15in
\begin{center}
\centering
    \begin{subfigure}[b]{0.21\textwidth}
        \centering
        \includegraphics[width=\textwidth]{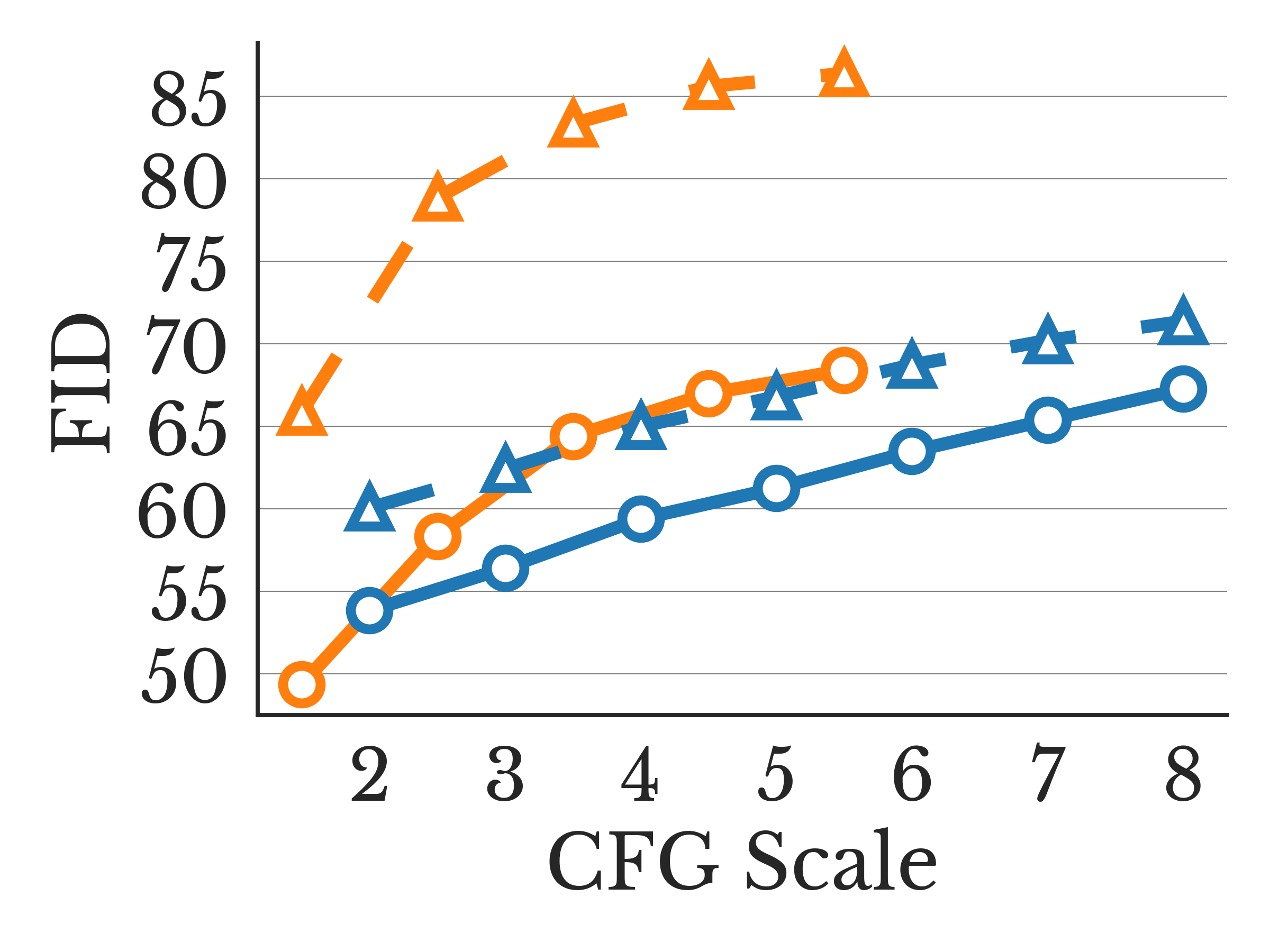}
    \end{subfigure}%
    \begin{subfigure}[b]{0.21\textwidth}
        \centering
        \includegraphics[width=\textwidth]{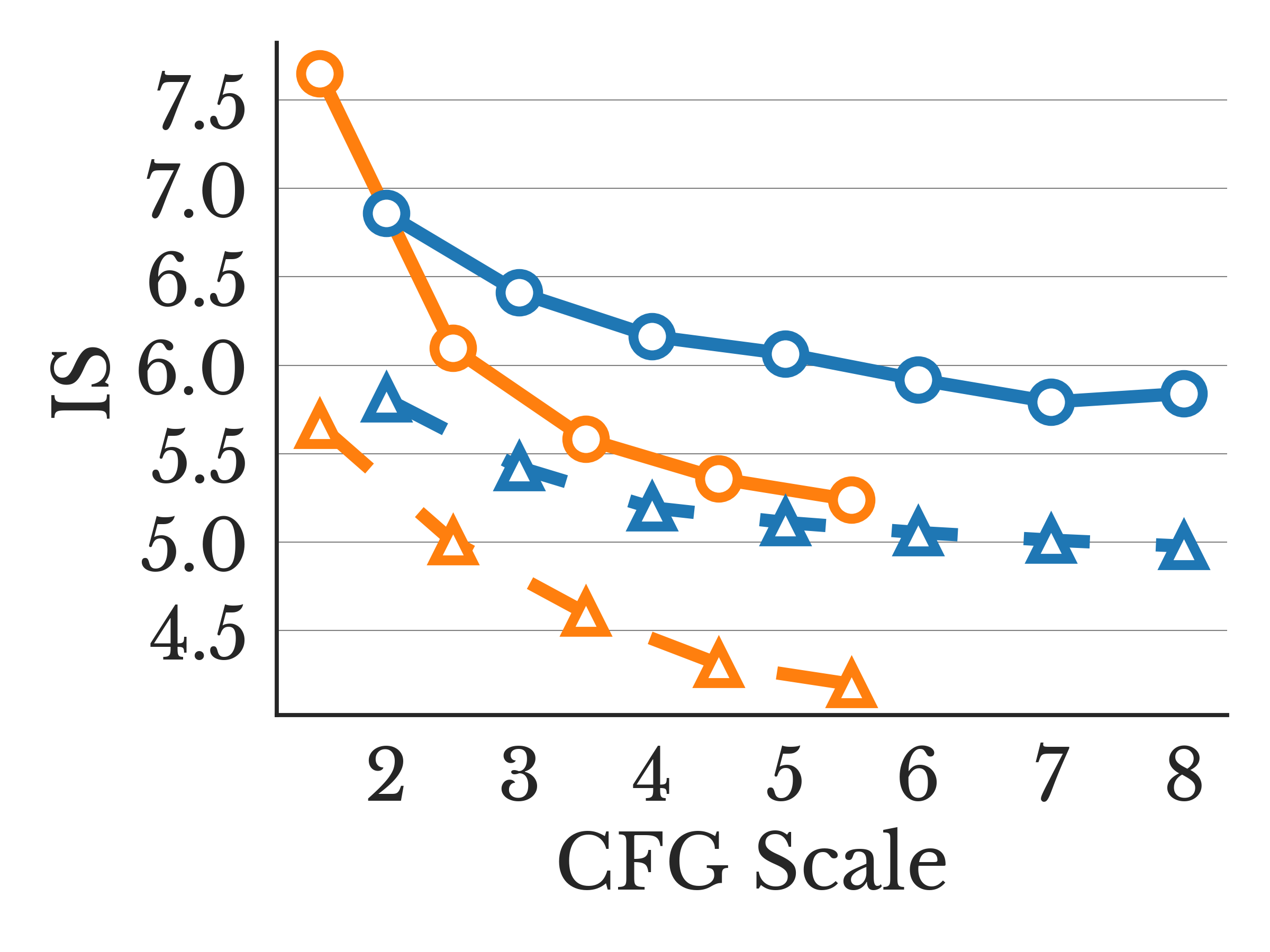}
    \end{subfigure}%
    \begin{subfigure}[b]{0.21\textwidth}
        \centering
        \includegraphics[width=\textwidth]{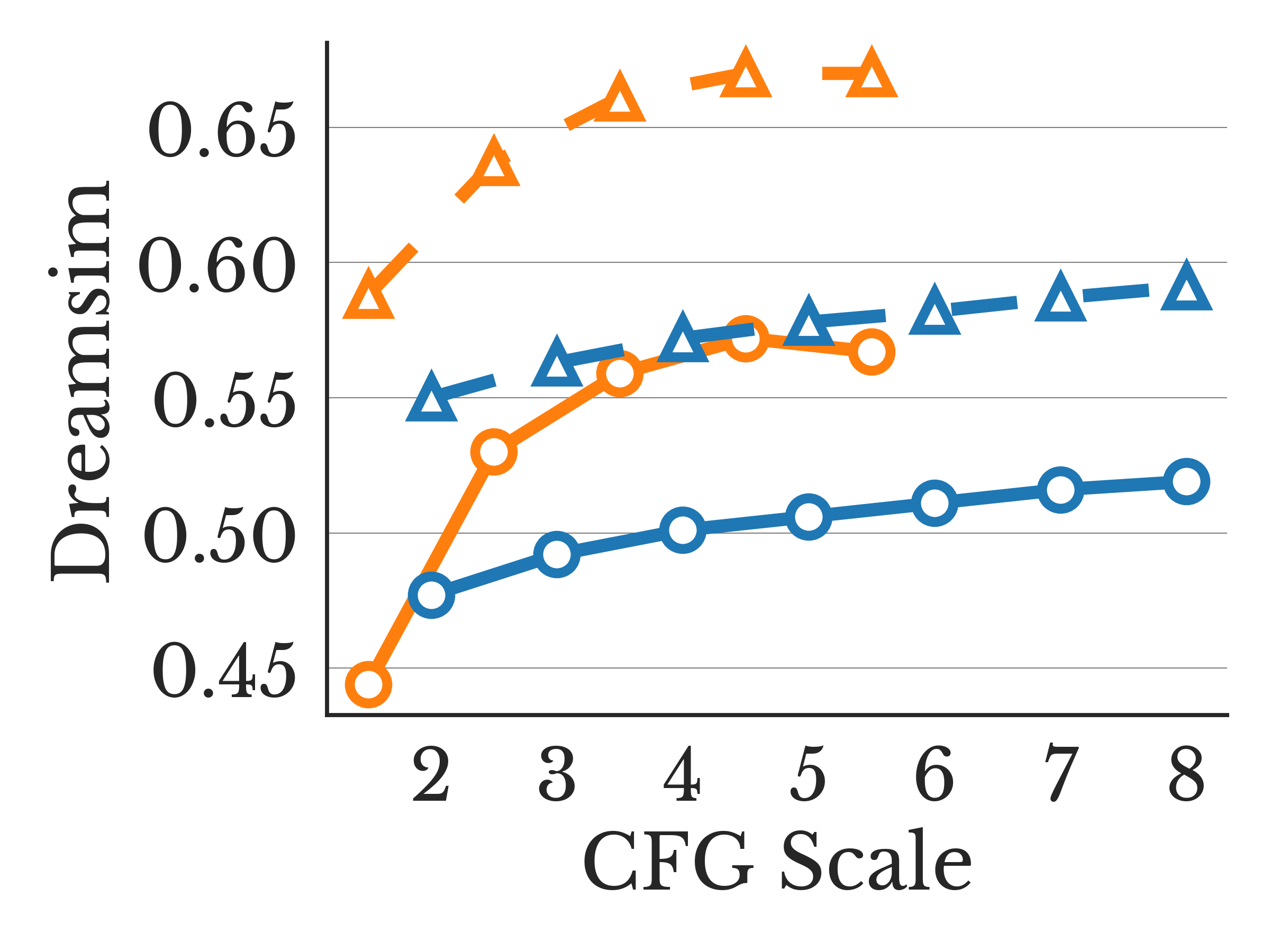}
    \end{subfigure}%
    \begin{subfigure}[b]{0.21\textwidth}
        \centering
        \includegraphics[width=\textwidth]{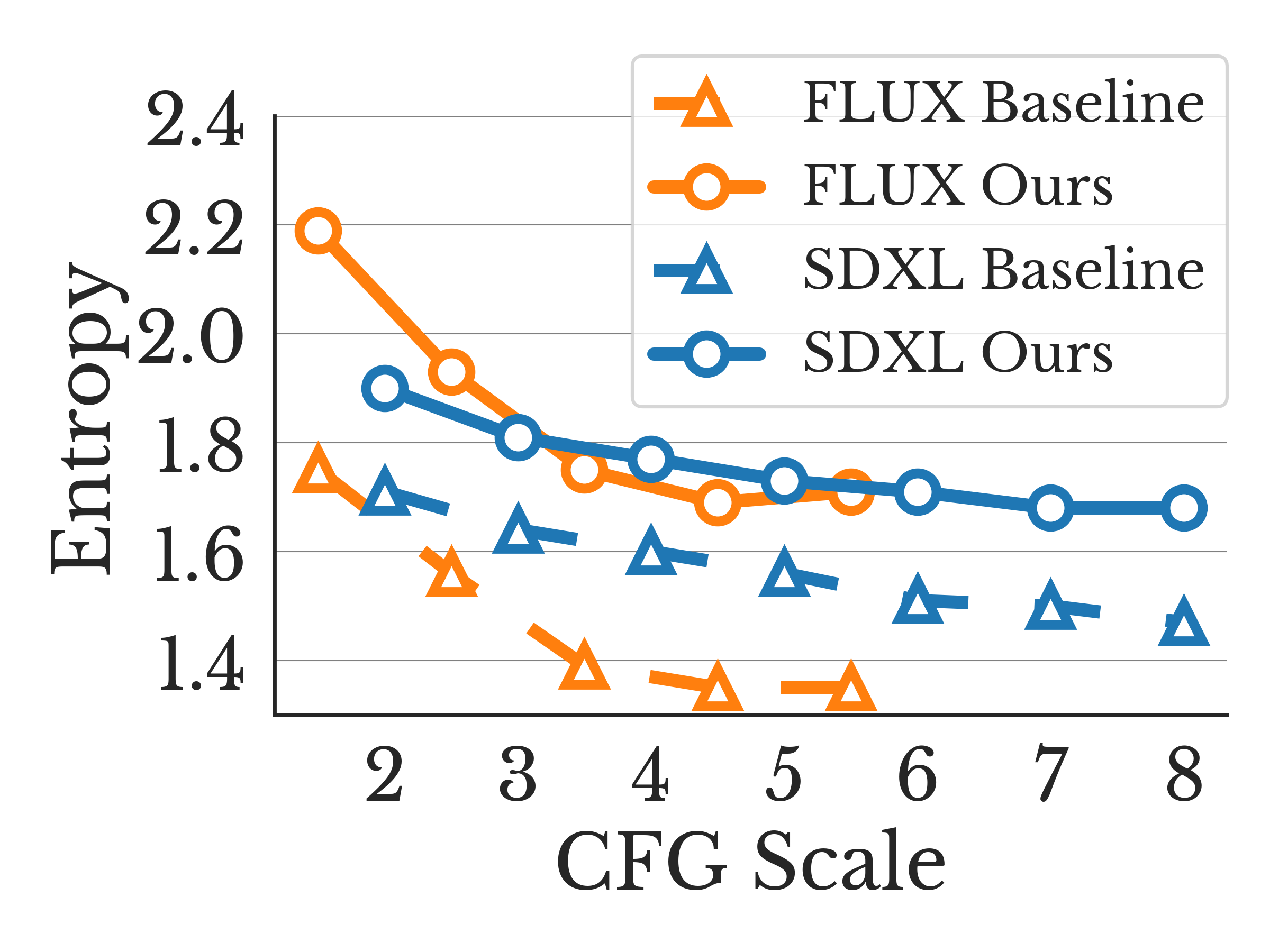}
    \end{subfigure}%
\vskip -0.15in
\caption{\emph{\textbf{Quantitative Results for Output Diversity}}. \negtome (ours) helps improve output diversity (lower DreamSim score and higher Entropy) while preserving or improving quality (lower FID and higher IS) across different CFG scales for both SDXL and FLUX.}
\label{fig:quant-results}
\end{center}
\vskip -0.2in
\end{figure*}

\begin{figure*}[t]
\vskip -0.1in
\begin{center}
\centerline{\includegraphics[width=0.95\linewidth]{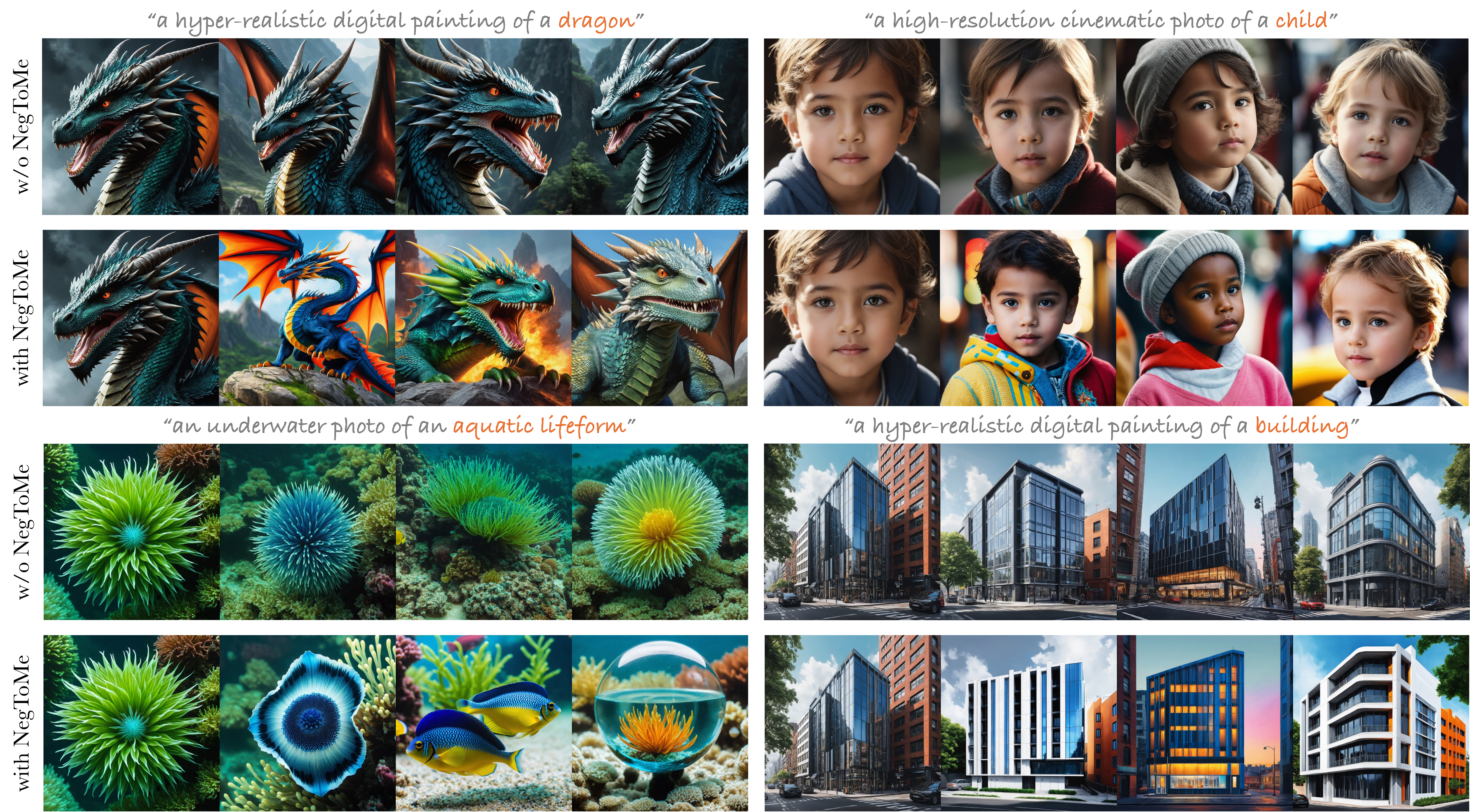}}
\vskip -0.1in
\caption{\emph{\textbf{Increasing Output Diversity.}} We observe that when performed \emph{w.r.t} to images in the same batch (the first image of each batch in above), \negtome significantly improves output diversity (racial, ethnic, visual) while preserving output image quality.}
\label{fig:diversity}
\end{center}
\vskip -0.35in
\end{figure*}

We evaluate the performance of our approach for increasing output diversity when performing negative token merging \emph{w.r.t} 
to other images in the batch. To facilitate easy visual comparison, we perform negative token merging \emph{w.r.t} 
the first image in each batch. Unless otherwise specified, all results are reported using the same text-to-image generation base-model \citep{podell2023sdxl,flux2024} on a single Nvidia-H100 GPU.

\noindent\emph{\textbf{Dataset and Setup.}} We first construct an input prompt dataset comprising 20 general object categories (\eg, animal, woman, bird, car \etc.) across 7 different prompt templates (\eg, ``a photo of a'', ``a high-contrast image of a'')\footnote{We adopt CLIP prompt templates~\cite{radford2021learning} for CIFAR10 image classification, excluding ones that imply low-quality (\eg, \emph{``a blurry photo of a''})}. For each category, we sample 280 images with 10 random seeds (4 per batch) both with and without \negtome. The real images for FID \cite{heusel2017gans} calculation are sourced from LAION-Aesthetics-v2 6+ dataset \cite{schuhmann2021laion}, where we use  CLIP~\cite{radford2021learning} to retrieve the top-1K images for each category.

\noindent\emph{\textbf{Evaluation metrics.}} 
Following \cite{ho2022classifier}, we report the results for output-quality using 1) FID \cite{heusel2017gans} and 2) Inception Score (IS) \cite{salimans2016improved}. 
3) Pairwise dreamsim-score \cite{fu2023dreamsim}: is used to measure output feature diversity. 4) VQAScore~\cite{lin2024evaluating} and 5) CLIPScore~\cite{hessel2021clipscore} are used to evaluate image-text alignment.
Furthermore following \cite{elgammal2017can}, we also use 6) Entropy-Score which measures the degree to which outputs for a particular object category (\eg, person) are spread across its subcategories (racial, gender, ethnic \etc). For human images, we use the FairFace classifier \cite{Krkkinen2019FairFaceFA} to detect race, gender, and age. For non-human categories (\eg, bird), sub-categories are extracted via WordNet \cite{miller1995wordnet} and classified using CLIP.

\begin{figure}[t]
\vskip -0.15in
\begin{center}
\centerline{\includegraphics[width=1.\linewidth]{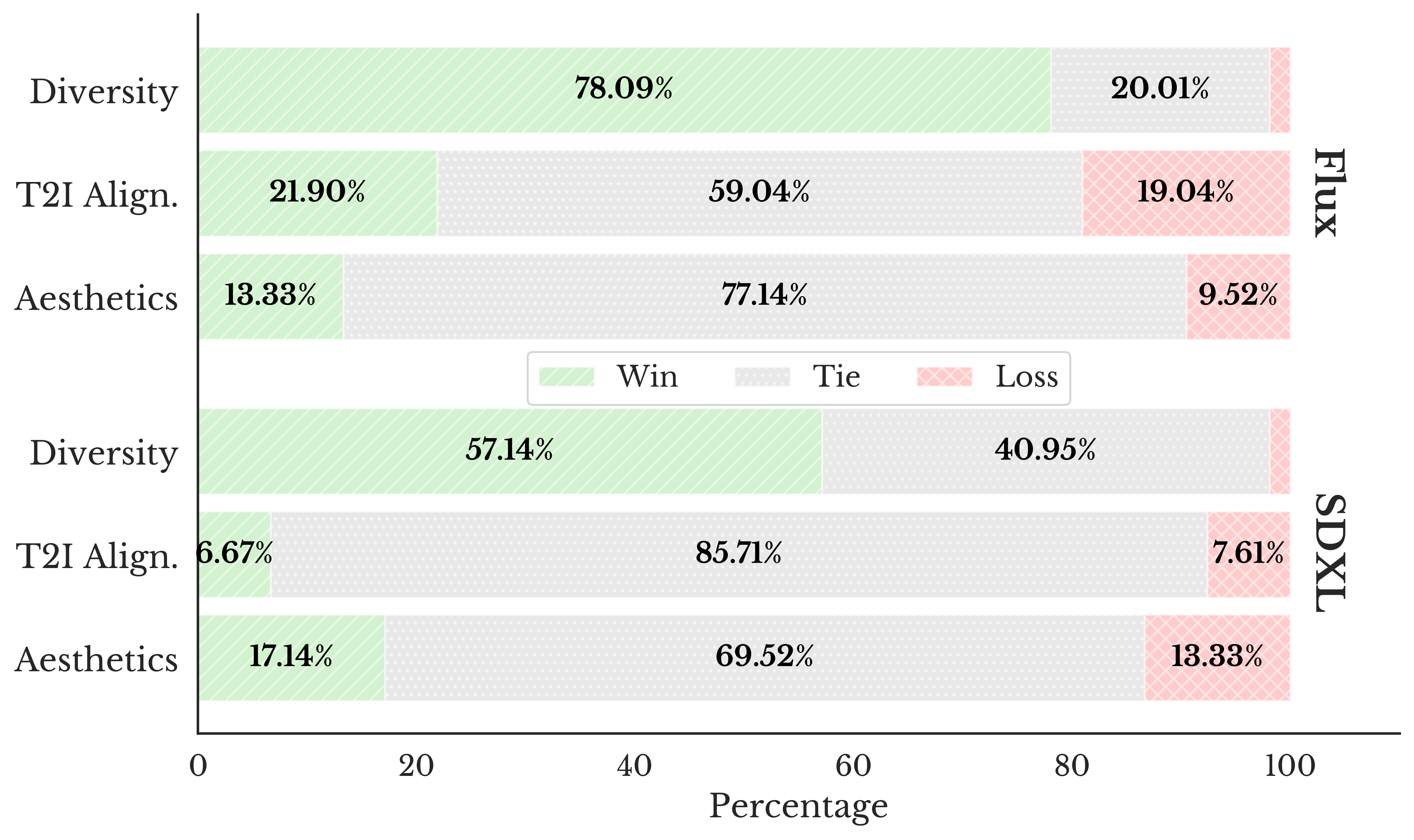}}
\vskip -0.1in
\caption{\emph{\textbf{Human User Study.}} \negtome helps improve output diversity while preserving text-to-image alignment performance.}
\label{fig:user-study}
\end{center}
\vskip -0.4in
\end{figure}

\noindent\emph{\textbf{Quantitative Results.}}
Results are shown in Figure~\ref{fig:quant-results}. We observe that \negtome helps improve output diversity (\ie lower Dreamsim scores and higher Entropy) while preserving or even improving quality (\ie lower FID and higher IS) across different classifier-free guidance (cfg) scales \cite{ho2022classifier} for both SDXL and FLUX. 
We also perform human evaluation 
evaluating diversity, quality and prompt alignment of the proposed approach using actual human users (Fig.~\ref{fig:user-study}). Similar to automated metric evaluation, we observe that NegToMe helps improve output diversity without sacrificing output image quality and prompt alignment performance.

\begin{figure}[t]
\vskip -0.15in
\centering
\begingroup
\setlength{\tabcolsep}{3.0pt}
\footnotesize
\begin{tabular}{l|cccc}
\toprule
Method & Dreamsim $\downarrow$ &  CLIPScore $\uparrow$ & IS $\uparrow$ & Inf. Time $\downarrow$\\
\hline
Base Prompt & 0.812  & 0.334 & 3.197 & 13.2 s \\
Base Prompt + Ours & 0.780 & \textbf{0.336} & 3.355 & 13.7 s\\
\hline
PW (\texttt{gpt-4o}) & 0.743 & 0.332 & 3.686 & 15.4 s\\
PW + Ours & \textbf{0.712} & 0.333 & \textbf{3.747} & 15.9 s \\
\bottomrule
\end{tabular}
\label{tab:prompt-rewriting}
\endgroup
\includegraphics[width=0.9\linewidth]{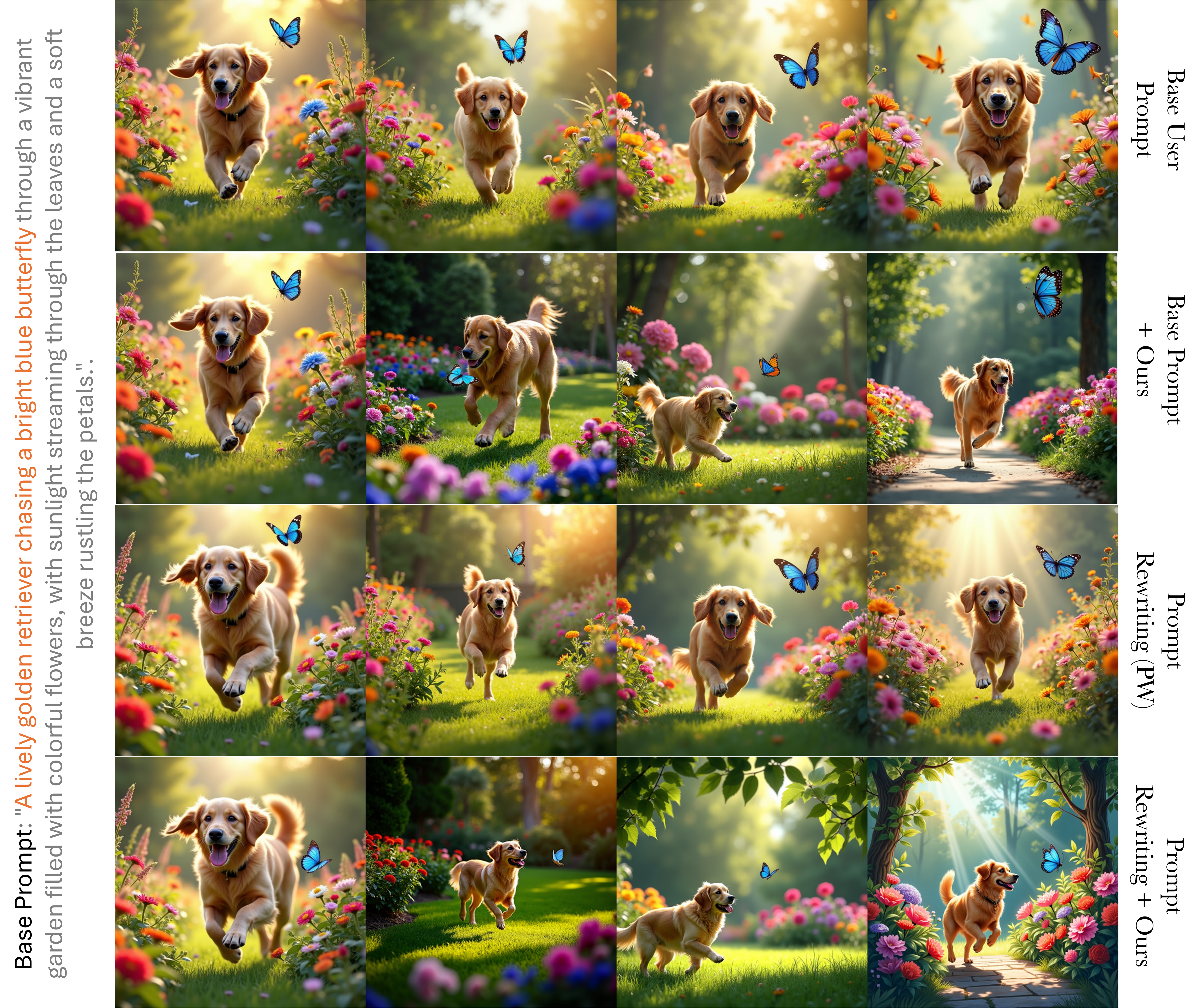}
\vskip -0.1in
\caption{NegToMe helps improve output diversity both with (row-2) and without explicit prompt-rewriting (PW) (row-4).}
\label{fig:prompt-rewriting}
\vskip -0.3in
\end{figure}

\begin{figure*}[t]
\vskip -0.2in
\begin{center}
\centerline{\includegraphics[width=0.95\linewidth]{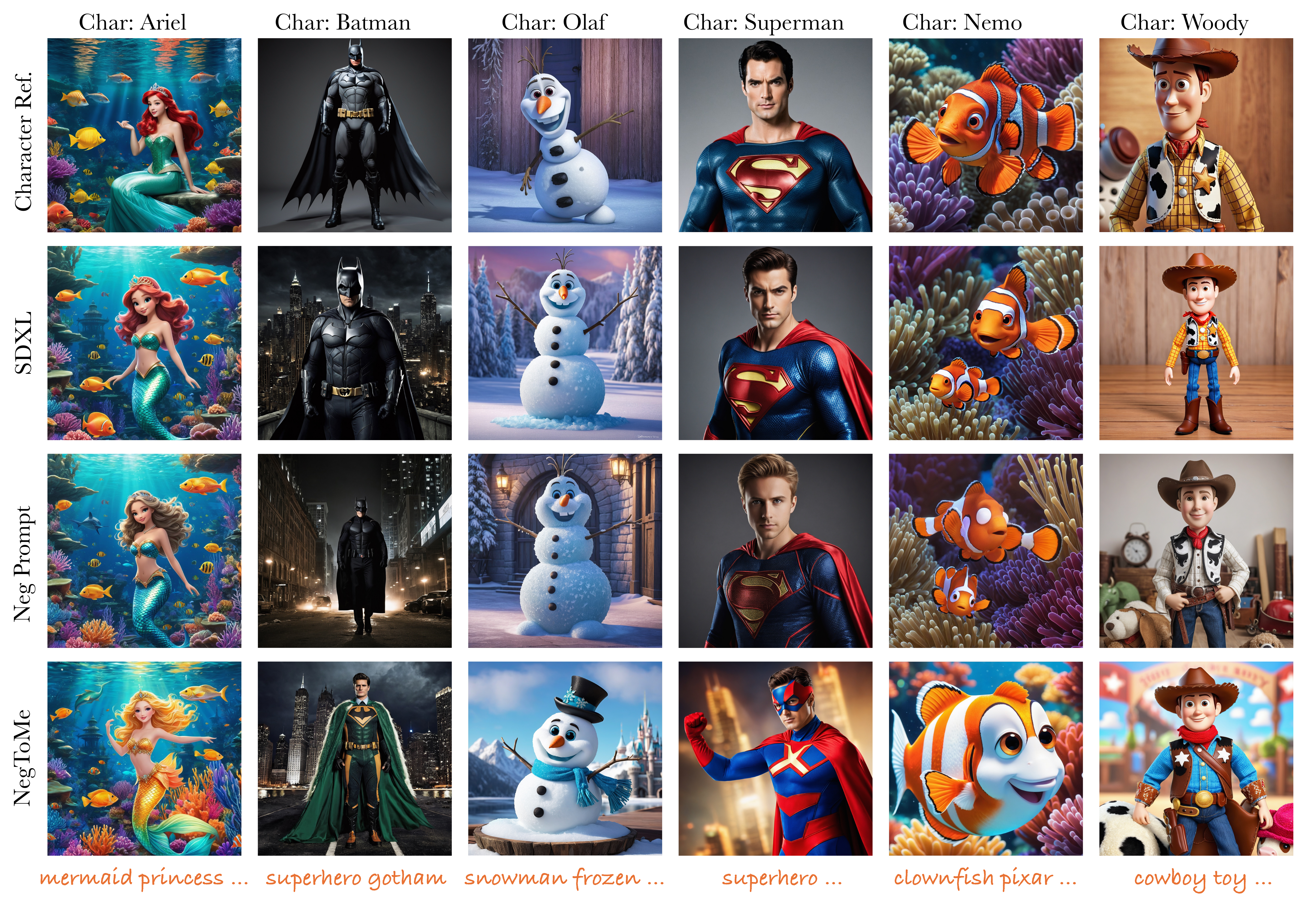}}
\vskip -0.2in
\caption{\emph{\textbf{Copyright Mitigation.}}  When used \emph{w.r.t} a copyright RAG image dataset, NegToMe helps reduce visual similarities with copyrighted characters without sacrificing output image quality (Tab.~\ref{tab:quant-results}). Complete prompts and further results are provided in the appendix.
}
\label{fig:copyright-qual}
\end{center}
\vskip -0.35in
\end{figure*}

\begin{figure*}[t]
\begin{center}
\centering
     \begin{subfigure}[b]{0.95\textwidth}
         \centering
         \includegraphics[width=1.\textwidth]{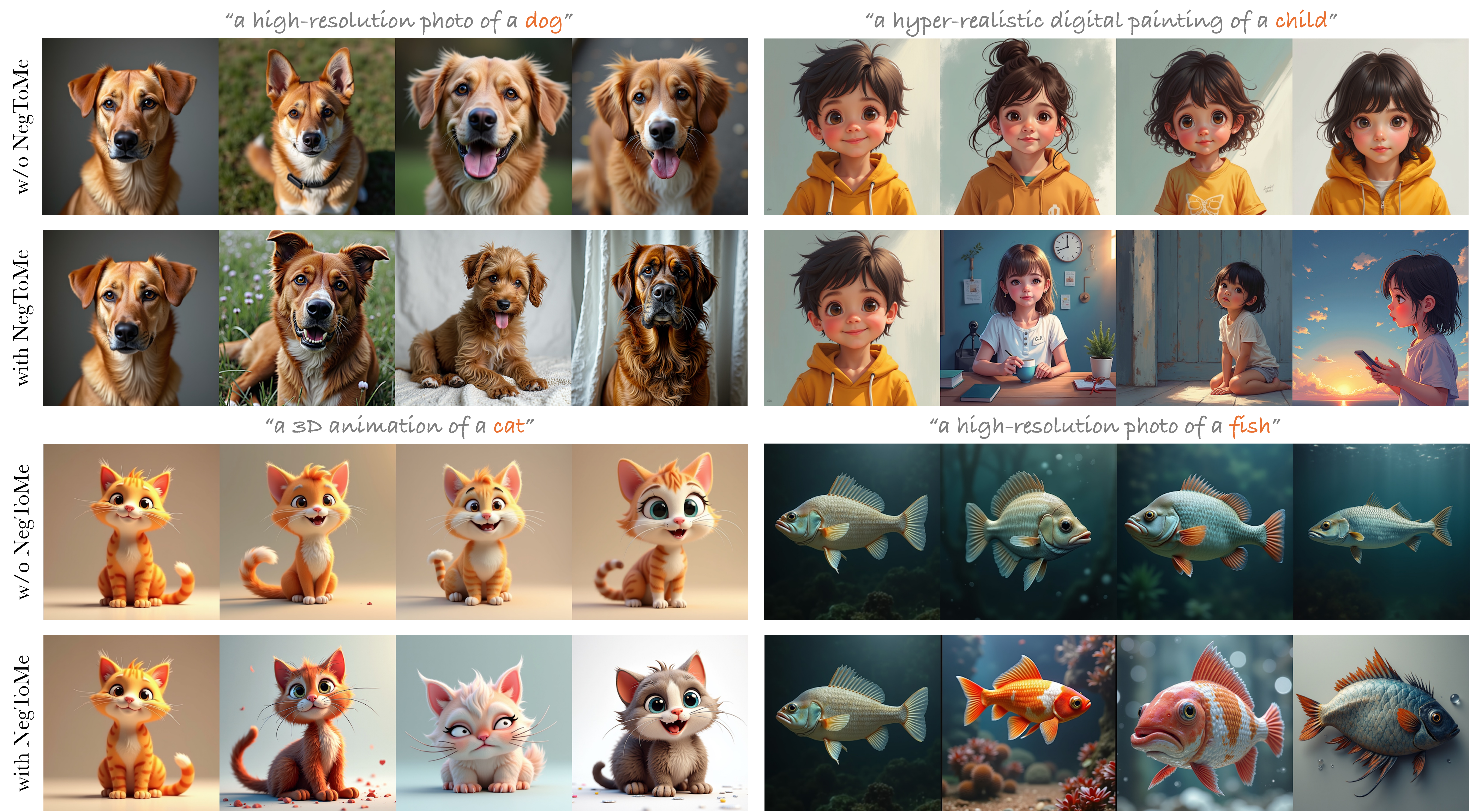}
     \end{subfigure}
     \begin{subfigure}[b]{0.95\textwidth}
         \centering
         \includegraphics[width=1.\textwidth]{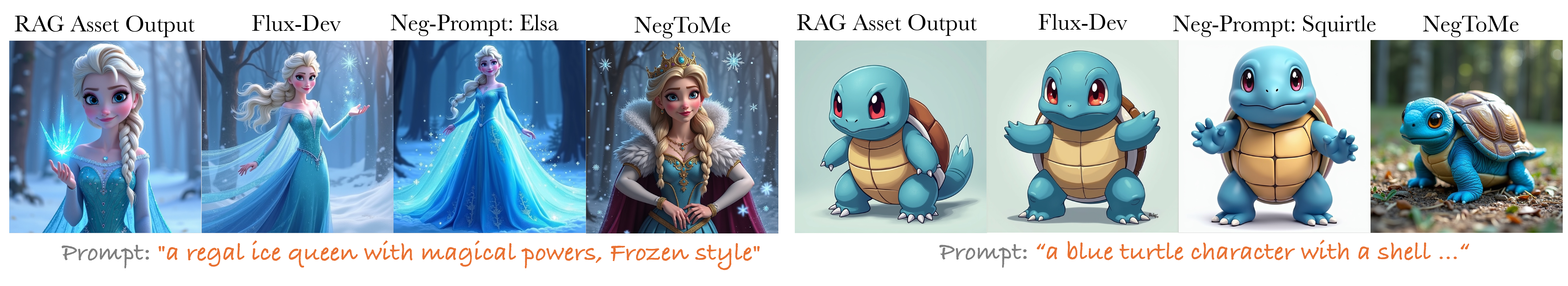}
     \end{subfigure}
\vskip -0.15in
\caption{\emph{\textbf{Application to MM-DiT models (Flux).}} \negtome is model-agnostic and  also applicable to MM-DiT models like Flux \cite{flux2024}. \negtome significantly increases the output diversity (\emph{top}), and helps reduce copyright violation (\emph{bottom}).}
\label{fig:mmdit}
\end{center}
\vskip -0.2in
\end{figure*}

\noindent\emph{\textbf{Qualitative Results.}}
We visualize the outputs with and without \negtome for the base model SDXL (Fig.~\ref{fig:overview} and~\ref{fig:diversity}) and FLUX (Fig.~\ref{fig:mmdit}). To avoid cherrypicking, all visualizations are from a fixed seed $0$. 
We observe that the base-model often suffers from the problem of mode-collapse with the generated images exhibiting limited visual feature diversity (\eg, visually similar dragons/buildings in Fig.~\ref{fig:diversity}, cats in Fig.~\ref{fig:mmdit}). Also the outputs might simply collapse into the same subcategory, even when the prompt is about a general category (\eg, aquatic lifeform in Fig.~\ref{fig:diversity}). In contrast, despite its simplicity, \negtome is able to better harness the underlying diffusion prior in order to generate diverse images in terms of demographics (\eg, person in Fig.~\ref{fig:overview}), subcategory (\eg, aquatic lifeform in Fig.~\ref{fig:diversity}), viewpoints (\eg, dragon in Fig.~\ref{fig:diversity}), image layout, pose, visual appearances of both foreground and background (\eg, child in Fig.~\ref{fig:mmdit}) \etc.

\begingroup
\setlength{\tabcolsep}{3.0pt}
\small
\begin{table}[t]
\begin{center}
\footnotesize
\begin{tabular}{cc|cccc}
\toprule
\multicolumn{2}{c|}{Mitigation Strategy}  & \multicolumn{4}{c}{Evaluation Metrics} \\
\hline
NegPrompt & NegToMe & Dreamsim $\downarrow$ & VQAScore $\uparrow$ & CLIPScore $\uparrow$ & IS $\uparrow$ \\
\hline
\xmark & \xmark & 0.766 & \textbf{0.913} & {0.344} & 3.431 \\
\cmark & \xmark & {0.684} & 0.876 & 0.339 & {3.790} \\
\hline
\xmark & \cmark & 0.703 & {0.906} & \textbf{0.346} & 3.678 \\
\cmark & \cmark & \textbf{0.638} & 0.879 & 0.339 & \textbf{3.864} \\
\bottomrule
\end{tabular}
\end{center}
\vskip -0.2in
\caption{
\emph{\textbf{Copyright Mitigation}}. 
\negtome reduces visual similarity to copyright characters while preserving  T2I performance.}
\label{tab:quant-results}
\vskip -0.25in
\end{table}
\endgroup

\noindent\emph{\textbf{Output diversity with explicit prompt-rewriting.}} 
A key advantage of \negtome is that it helps improve diversity without the need for extensive prompt-rewriting, which also presents as a feasible yet expensive (time \& memory) approach for improving diversity. This is particularly relevant when the user-prompts are quite detailed and long. 
To provide a fair comparison with prompt-rewriting setting, we therefore first curate a set of 20 detailed prompts across diverse settings (see appendix). For each prompt we then use a large-language model \cite{achiam2023gpt} in order to generate diverse variations of the original base prompt. The final images for both base-prompt and rewritten prompts are sampled across 10 random seeds.
Results are shown in Fig.~\ref{fig:prompt-rewriting}.
While prompt-rewriting helps improve diversity, it comes at the cost of increased inference time. Furthermore, some of the generated outputs might still appear similar (\eg, col-1 and col-3: Fig.~\ref{fig:prompt-rewriting}). In contrast, \negtome can adaptively improve the output diversity (with both base and rewritten prompts), while on average using only $<4\%$ higher inferences times.

\subsection{Copyright Mitigation}
\label{sec:copyright}

We next show the efficacy of our approach for reducing visual similarities with copyrighted characters when performing \negtome \emph{w.r.t} a copyrighted image RAG database. 

\noindent\emph{\textbf{Dataset and Setup.}} 
We first construct a dataset of 50 copyrighted characters (\eg, Mario, Elsa, Batman), and curate input prompts which to trigger these characters
without explicitly mentioning their names (see appendix). For each character, we compile a reference dataset of approximately 30 high-quality images depicting the character in diverse settings. Masked negative token merging is then performed for each prompt, using the best-matching RAG asset (asset with highest Dreamsim score) from the reference dataset.
The mask for each asset is computed using HQ-SAM \cite{sam_hq}.

\noindent\emph{\textbf{Qualitative Results.}} Results are shown in Fig.~\ref{fig:copyright-qual} (SDXL) and Fig.~\ref{fig:mmdit} (Flux). 
We observe that the base model still generates copyrighted characters, even when the corresponding character name is not mentioned in the input prompt.
Using the character name as the negative prompt alone often is not sufficient, as the output images still show high visual similarity to the copyrighted character. In contrast, by applying adversarial guidance directly using character visual features, \negtome reduces similarity to copyrighted characters while maintaining text-to-image alignment.

\begin{figure}[t]
\vskip -0.15in
\centering
\begingroup
\setlength{\tabcolsep}{3.0pt}
\footnotesize
\begin{tabular}{l|cccc}
\toprule
Method & AES $\uparrow$ & VQAScore $\uparrow$ & CLIPScore $\uparrow$ & Human Pref. $\uparrow$ \\
\hline
Flux-Dev & 6.428 & \textbf{0.866} & 0.320 & 22.5 \% \\
Flux-Dev + Ours & \textbf{6.604} & 0.861 & \textbf{0.322} & \textbf{77.5} \% \\
\bottomrule
\end{tabular}
\endgroup
\includegraphics[width=1.\linewidth]{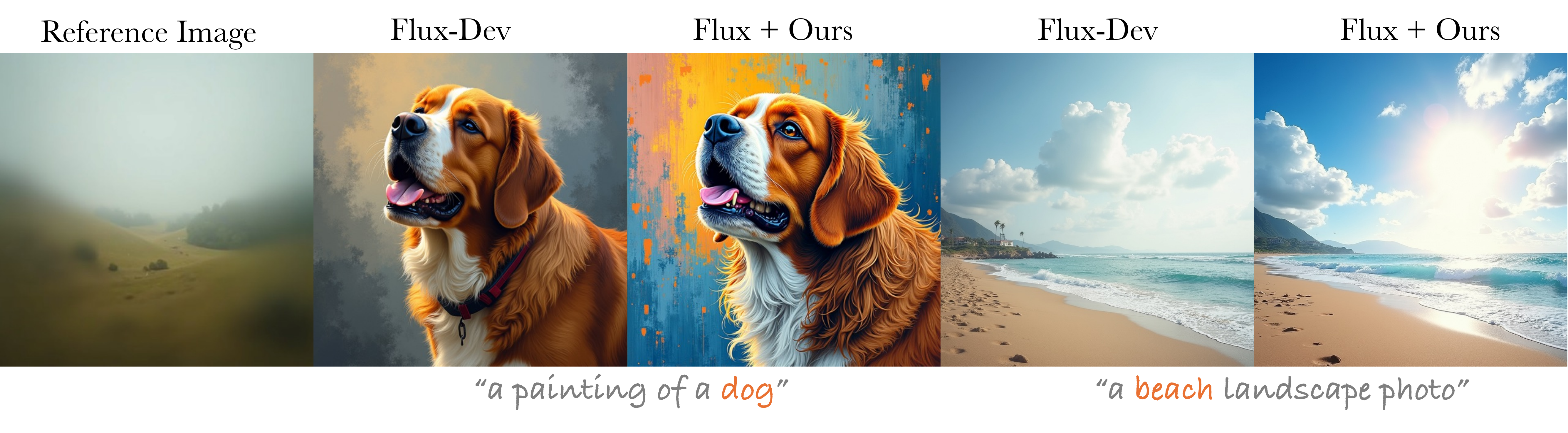}
\vskip -0.15in
\caption{\emph{\textbf{Improving aesthetics.}} Using a blurry reference with \negtome improves output aesthetics without any training \cite{rafailov2024direct}.}
\label{fig:output-aesthetics}
\vskip -0.2in
\end{figure}

\begin{figure*}[t]
\vskip -0.2in
\begin{center}
\centerline{\includegraphics[width=0.98\linewidth]{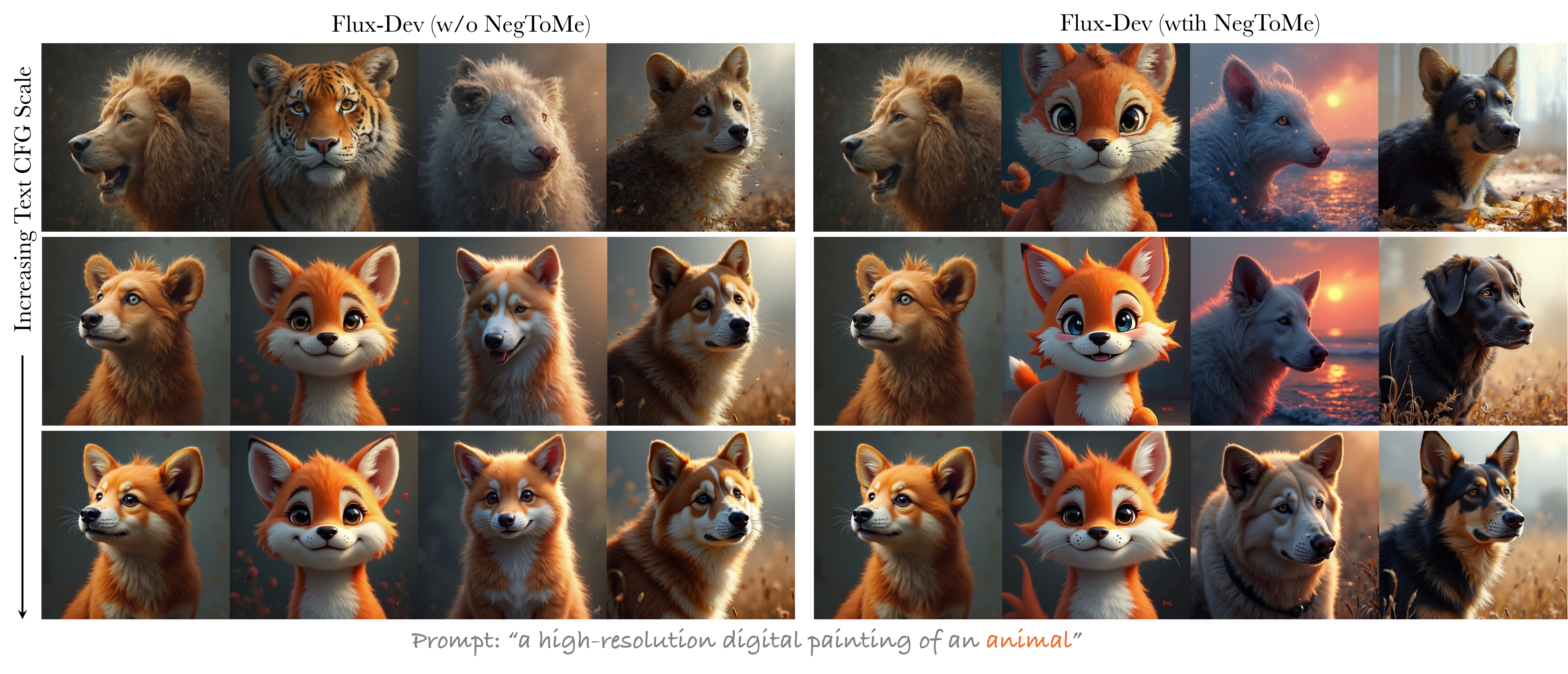}}
\vskip -0.2in
\caption{\emph{\textbf{Variation with cfg scale}} 
leads to improved output quality at the cost reduced diversity (\emph{left}). NegToMe not only improves output quality at lower \emph{cfg} values (by guiding away from poor-quality features, see Fig.~\ref{fig:output-aesthetics}) but also helps improve output diversity for higher \emph{cfg}.}
\label{fig:cfg-var}
\end{center}
\vskip -0.3in
\end{figure*}

\begin{figure}[t]
\vskip -0.15in
\begin{center}
\centerline{\includegraphics[width=1.\linewidth]{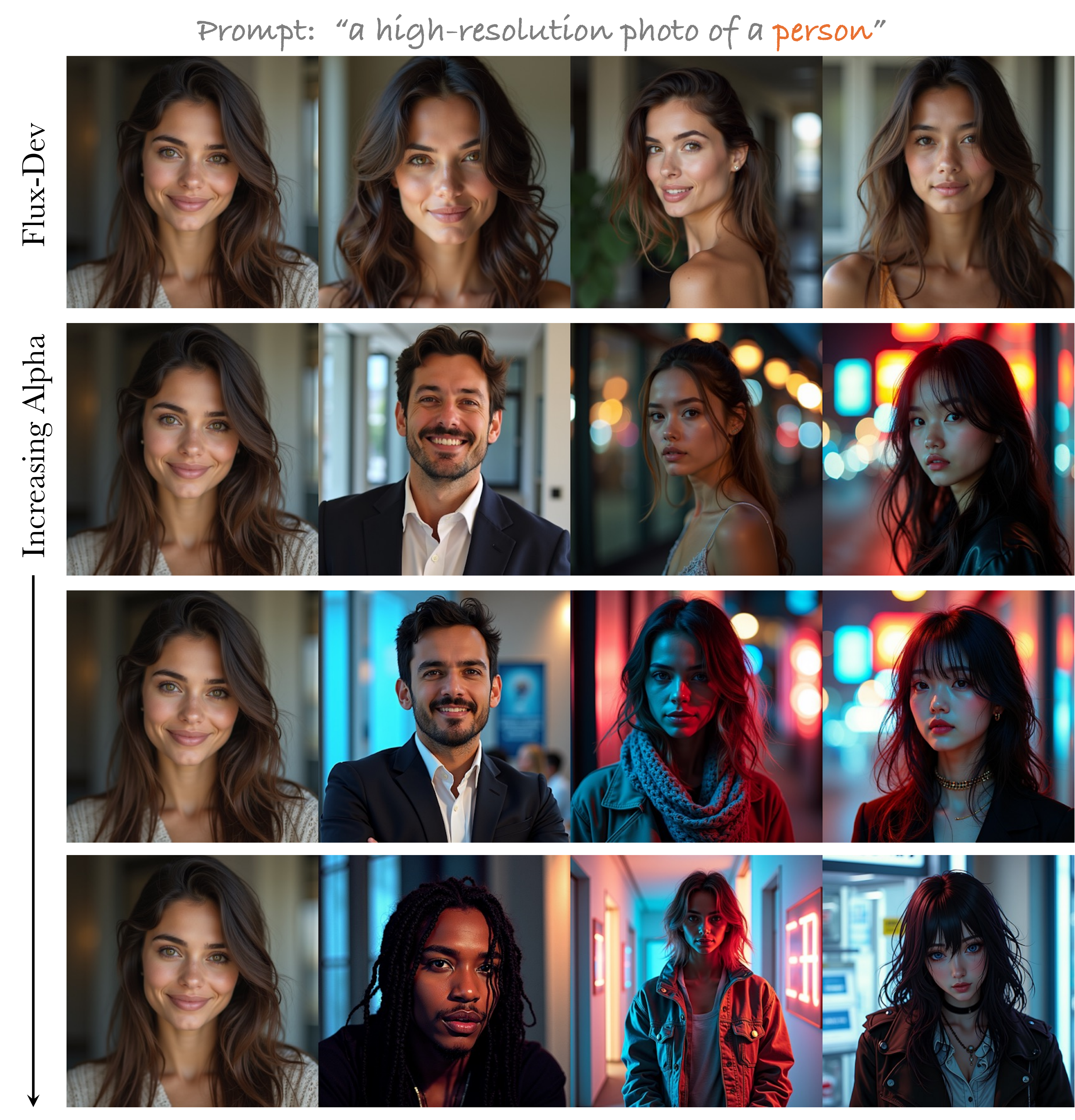}}
\vskip -0.1in
\caption{\emph{\textbf{Variation with merging alpha.}} Increasing the value of $\alpha$ (refer to Sec.~\ref{sec:method}) for \negtome gradually increases output diversity in terms of gender, race, ethnicity, lighting, style \etc.}
\label{fig:alpha-var}
\end{center}
\vskip -0.35in
\end{figure}

\noindent\emph{\textbf{Quantitative Results.}}
We evaluate our approach using the base model SDXL, comparing it to pure negative-prompt copyright mitigation strategies. For each prompt, we sample 50 images with 50 different random seeds. For evaluation, we use the 1) maximum DreamSim score \cite{fu2023dreamsim} across all RAG assets (excluding the reference used for \negtome) for measuring visual similarity to copyrighted characters.  2) VQAScore \cite{lin2024evaluating}, CLIPScore \cite{hessel2021clipscore} for text-to-image alignment, 3) IS \cite{heusel2017gans} for image quality.
Results are shown in Tab.~\ref{tab:quant-results}. We observe that \negtome reduces visual similarity to copyrighted characters without sacrificing text-to-image alignment and image quality. Furthermore, \negtome is  complementary to negative prompting, with the best performance achieved when combining both methods.

\section{Method Analysis and Applications}
\label{sec:analysis}

\emph{\textbf{Improving output aesthetics.}}
As noted in Fig.~\ref{fig:general-use}, we note that \negtome allows for a range of custom applications by appropriately adjusting the reference inputs. Notably, we find that when using a poor quality image as reference, \negtome helps improve output aesthetics and image quality without requiring any training / finetuning \cite{rafailov2024direct} (Fig.~\ref{fig:output-aesthetics}).

\emph{\textbf{Variation across text-guidance scale.}}
Results are shown in Fig.~\ref{fig:cfg-var}. We observe that traditional text-based classifier-free guidance \cite{ho2022classifier} suffers from a tradeoff between output diversity and image quality. In contrast, we find that NegToMe is able to improve output diversity while preserving image quality across different scales of classifier-free guidance. Interestingly, we also observe that the increase in output diversity with NegToMe is often accompanied by an increase in output image quality especially at lower \emph{cfg} values. This happens due to the use of a poor quality reference image (\eg, \emph{lion}: Fig.~\ref{fig:cfg-var}) at lower \emph{cfg} scales, which in addition to improving diversity also tends to improve output image aesthetics and details (\emph{lamb, dog} in row-1: Fig.~\ref{fig:cfg-var}).

\emph{\textbf{Variation with merging alpha.}} 
Results are shown in Fig.~\ref{fig:alpha-var}. We observe that NegToMe provides an easy to use mechanism for controlling  output image diversity. As seen in Fig.~\ref{fig:alpha-var}, we observe that gradually increasing the value of $\alpha$ (Sec.~\ref{sec:method}) helps the user easily control output diversity in terms of race, gender, ethnicity, lighting, style \etc.

\section{Conclusion}

In this paper we introduce \negtome, a simple training-free approach which complements traditional text-based negative-prompt guidance, by performing adversarial guidance directly using visual features of a reference image. \negtome is simple, training-free and can be incorporated with most \emph{state-of-art} diffusion models using just few lines of code (Alg.~\ref{alg:negtome}). By simply varying the reference image, NegToMe enables a range of custom applications such as increasing output diversity (Sec.~\ref{sec:output-diversity}), reducing similarity with copyrighted images (Sec.~\ref{sec:copyright}), improving output aesthetics (Fig.~\ref{fig:output-aesthetics}) \etc, while on average using only marginally $<4\%$ higher inference times. We excitedly hope that our research helps users better leverage \emph{state-of-the-art} diffusion models for diverse creative applications.

\section*{Acknowledgments}
We would like to thank Ishan Misra for helpful discussions and feedback on experiment design and quantitative evaluations. We are also thankful to Jonas Kohler and Junshen Chen for early discussions on negative token merging.

{\small
\bibliographystyle{ieee_fullname}
\bibliography{main}
}


\end{document}